\documentclass[11pt,a4paper]{article}
\usepackage[hyperref]{acl2021}
\usepackage{times}
\usepackage{latexsym}

\usepackage{graphicx}
\usepackage{threeparttable}

\usepackage{diagbox}
\usepackage{amsmath}
\usepackage{array}
\usepackage{amsfonts}
\usepackage{amssymb}
\usepackage{multirow}
\usepackage{pifont}
\newcommand{\xmark}{\ding{55}}%
\newcolumntype{M}[1]{>{\centering\arraybackslash}m{#1}}
\newcolumntype{A}[2]{%
    >{\minipage{\dimexpr#1\linewidth-2\tabcolsep-#2\arrayrulewidth\relax}\vspace\tabcolsep}%
    c<{\vspace\tabcolsep\endminipage}}
\usepackage{microtype}

\aclfinalcopy 

\title{MCL@IITK at SemEval-2021 Task 2: Multilingual and Cross-lingual Word-in-Context Disambiguation using Augmented Data, Signals, and Transformers}

\author{
    Rohan Gupta \qquad   
    Jay Mundra$^{*}$ \qquad 
    Deepak Mahajan\thanks{\quad Authors equally contributed  to this work.} \qquad 
  \large{\textbf{Ashutosh Modi}} \\
{Indian Institute of Technology Kanpur (IIT Kanpur)} \\
{\tt \{rohangpt, jaym, dipakam\}@iitk.ac.in}\\
  {\tt ashutoshm@cse.iitk.ac.in}  \\
}

\date{}

\begin{document}
\maketitle
\begin{abstract}

In this work, we present our approach for solving the SemEval 2021 Task 2: Multilingual and Cross-lingual Word-in-Context Disambiguation (MCL-WiC). The task is a sentence pair classification problem where the goal is to detect whether a given word common to both the sentences evokes the same meaning. We submit systems for both the settings - Multilingual (the pair's sentences belong to the same language) and Cross-Lingual (the pair's sentences belong to different languages). The training data is provided only in English. Consequently, we employ cross-lingual transfer techniques. Our approach employs fine-tuning pre-trained transformer-based language models, like ELECTRA and ALBERT, for the English task and XLM-R for all other tasks. To improve these systems' performance, we propose adding a signal to the word to be disambiguated and augmenting our data by sentence pair reversal. We further augment the dataset provided to us with WiC, XL-WiC and SemCor 3.0.  Using ensembles, we achieve strong performance in the Multilingual task, placing first in the EN-EN and FR-FR sub-tasks. For the Cross-Lingual setting, we employed translate-test methods and a zero-shot method, using our multilingual models, with the latter performing slightly better.

\end{abstract}

\section{Introduction}
A key challenge in lexical semantics is to identify or to encode the different senses of an ambiguous word.  
The Word Sense Disambiguation task (WSD) \cite{wsd} is a framework used to evaluate systems in their ability to identify different senses of the word. The task involves selecting the correct sense (meaning) of a target word from a list of senses listed in a sense inventory like WordNet \cite{fellbaum2012wordnet}. \citet{pilehvarwic} proposed a novel benchmark (WiC - Word in Context Disambiguation) for the task casting the problem as a binary classification task, wherein it has to be identified whether a word common to a sentence pair is used in the same sense or not. The WiC task frees up the word sense disambiguation task from being tied to any sense inventory.

\begin{figure}[h!]
\includegraphics[scale = 0.25]{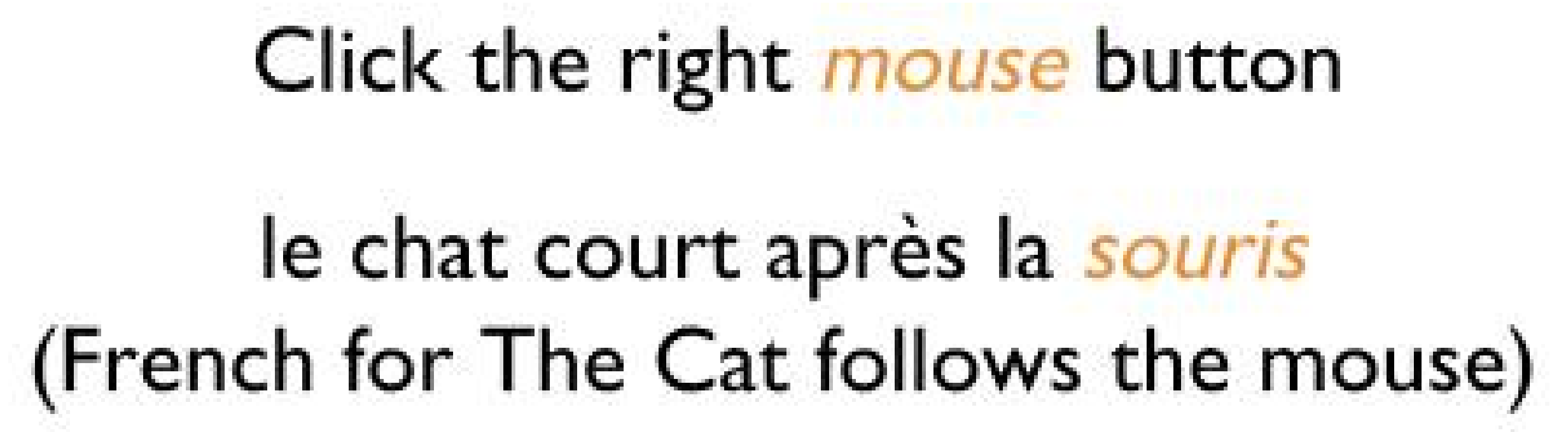}
\caption{An demonstrative example for the English-French Cross-lingual dataset. This pair will be classified as a 'False' pair.}
\label{fig:example}
\end{figure}

The SemEval 2021 Task 2: Multilingual and Cross-lingual Word-in-Context Disambiguation \cite{martelli-etal-2021-mclwic} extends the WiC framework proposed by \citet{pilehvarwic} to more languages. The task is divided into two subtasks - the Multilingual task and the Cross-Lingual task. The sentence pair, with a word in common, which is to be disambiguated, is drawn from the same language in the MultiLingual task, whereas the pair is drawn from two different languages in the Cross-Lingual Task. The task is posed as binary classification task over a pair of sentence wide contexts $sent1$ and $sent2$, containing word sequences $w_1$ and $w_2$ respectively. The word sequences, $w_1$ and $w_2$, have a common word in lemmatized form $lemma$. When $w_1$ and $w_2$ invoke the same sense of the lemma in their respective contexts, it is to be labeled as `T' (True) class, else it labeled 'F' (False).  
As mentioned before, for the Multilingual setting, $sent1$ and $sent2$ are from the same language; for the Cross-lingual setting, $sent1$ is from English, and $sent2$ is from a non-English language. The languages considered for the Task are Arabic (AR),  English (EN), French (FR), Russian (RU), and Chinese (ZH). Therefore for the Multilingual evaluation, we have AR-AR, EN-EN, FR-FR, RU-RU, and ZH-ZH settings, and for the Cross-Lingual evaluation, we have EN-AR, EN-FR, EN-RU, and EN-ZH settings. An example of Cross-Lingual sentence pair is given in figure \ref{fig:example}. This task provides an evaluation benchmark for word sense disambiguation systems in languages other than English, a direction that has been less explored.

For this task, the training data is only provided for the EN-EN setting, and the development sets are provided only for the Multilingual task. As we are proposing supervised systems, it is essential to consider if we have training data or not. Therefore, splitting the Multilingual task, we propose systems for three components - (i) EN-EN (train and dev data available), (ii) Non-English Multilingual (only dev data available), and (iii) Cross-Lingual (neither train nor dev data available).
Our models and implementations are available here\footnote{https://github.com/dipakamiitk/Crosslingual-WSD.git}.

\section{Related Work}
\noindent\textbf{Word Sense Disambiguation: } The techniques for the WSD task are broadly divided into knowledge-based and supervised approaches. The supervised approaches include fine-tuning BERT for sequence classification \cite{wang2019superglue}, EWISE \cite{ewise}, and BiLSTM with attention \cite{raganato-etal-2017-neural}. The knowledge-based methods use the information present in sense inventories such as WordNet  \cite{fellbaum2012wordnet}, BabelNet  \cite{navigli2012babelnet} and Wikipedia to derive semantic knowledge, assisting in the task of WSD. These include building sense embeddings for each sense of the word and disambiguate the target word using the nearest neighbour sense embedding: SensEmBERT \cite{sensembert}, \cite{lmms}, or augmenting the pretraining objecting of BERT to take into account the sense information available in the WordNet: SenseBERT \cite{sensebert}. 

There are two existing benchmarks to evaluate the performance of WSD systems. One method is linked to the sense inventories, and the task is framed as a multi-class classification among the senses of a word listed in the inventory \cite{raganato-etal-2017-word}. The other is the WiC framework, not tied to any sense inventory, and asks if a target word has the same sense or not in the two given sentences. Recently, transformer-based architectures\cite{attention} (e.g., T5 \cite{t5}) have outperformed all existing approaches when fine-tuned on the WiC task. 

\noindent\textbf{Cross-Lingual NLP:} There are many NLP tasks for which the data is present in only some high resource language (often English), but the task needs to be solved for other languages. Some existing methods involve - (i) using multi-lingual language models (like mBERT \cite{bert}, XLM-R \cite{xlmr}) to train on the high resource language and transfer the learning to other languages, or (ii)  the translation approaches where we can translate train or test data to a target language and train a language-specific model. These methods have shown good performance on benchmarks like XGLUE \cite{liang2020xglue} and XTREME \cite{hu2020xtreme}, which have been designed to test this cross-lingual transfer performance of systems, with train data being present majorly in English, while the testing is to be done in other languages.

\section{Corpus Description and Data Augmentation}
\begin{table}[t]
\centering
\begin{center}

    \begin{tabular}{|M{1cm}|M{1.9cm}|M{1.9cm}|M{1.2cm}|}
    \hline
          & Multilingual (EN-EN) & Multilingual (Others) & Cross-Lingual \\ \hline
       
        Train Data &{ 8000} & { \xmark } & { \xmark } \\ \hline
        
        Dev Data & 1000 & 1000 (each) &  { \xmark }\\ \hline

    \end{tabular}
\caption{Available data for this task. All datasets are balanced, that is, equal number of `True' and `False' pairs.}
\label{dataset:stats}
\end{center}
\end{table} 
\label{sec:data}
A brief summary of the available data is shown in Table \ref{dataset:stats}. This data is manually curated and covers four parts of speech - Nouns, Verbs, Adjectives, and Adverbs. In addition, we augmented the data by utilizing WiC, XL-WiC, and SemCor.

\noindent\textbf{WiC:} We use the data provided by the WiC task \cite{pilehvarwic}, which proposes the same problem as this task but only in English. They collected their data semi-automatically from WordNet \cite{fellbaum2012wordnet}, and covered only Nouns and Verbs.

\noindent\textbf{XL-WiC: } It is a dataset that is an extension of the WiC dataset to multiple languages (equivalent to our multilingual setting) \cite{xlwic} . Again, the data was collected automatically from WordNet and Wiktionaries of various languages. We are only interested in a few languages, from the WordNet development sets, the ones with good human performance, because we often note that this data does not accurately represent human distinguishable senses, which our manually collected task data set does. Specifically, we use Chinese(ZH), Danish(DA), Croatian(HR), and Dutch(NL). Farsi(FA) also has good human performance but we could not include it due to a pre-processing error.

\noindent\textbf{AuSemCor:} We created our own augmented dataset AuSemCor from the SemCor \cite{miller1993semantic} dataset, which is a sense annotated corpora in English, with senses tagged using WordNet as its sense inventory. To generate data points for the same sense (T class), we pair up sentences containing a common lemma, whose WordNet senses are identical. For the other class (F class), we pair up sentences with a common lemma, but this lemma has different WordNet sense across the sentence pair. In addition, for the F class, we make sure that the WordNet supersense is also different for creating coarser sense distinctions as suggested by \citet{pilehvarwic}. We obtain 4986 datapoints with 2520 unique words. It is approximately balanced (2495 `F' and 2491 `T' pairs). Like WiC, it covers only Nouns and Verbs.

\section{Proposed Approach}
We shall now describe our proposed approaches for the two subtasks, multilingual and cross-lingual. We deal with English separately because training data is available in English and not in any other language. Since the data across the two tasks differ only in the language pairs, some general approaches apply to both settings. We finally submit an ensemble of models in all the tasks. The ensembling was done by taking average of the probability scores of the models (Probability Sum Ensemble).

\subsection{Task Agnostic Proposals}
\noindent \textbf{Signals: } \label{taskagnostic} We use a data preprocessing step of applying a signal to indicate the word to be disambiguated. This can be done in two ways - (i) \textit{Signal 1}: encoding the target word (the word to be disambiguated) in both the sentences of a pair within double quotes (e.g., \textit{Click the right `` mouse " button}) as suggested by \citet{glossbert}, or (ii) \textit{Signal 2}: append the target word at the end of the second sentence, similar to what was done by \citet{wang2019superglue}. Note, for the former method; we need the character spans of the target word to apply double quotes at the correct position.

\noindent\textbf{Sentence Reversal Augmentation: } For the models proposed in this task, the sentence pair is fed to the model in a manner such that the results do depend on which order the sentences are fed (i.e. the network parameters are not symmetric with respect to the two sentences). In such a case we propose the following augmentation - for every data point \textit{(sent1, sent2, lemma, label)}, add another data point \textit{(sent2, sent1, lemma, label)} to the set of data points. A similar notion can be extended to making more robust predictions - at inference, before thresholding on the probability scores returned by the model, take into account the reversed sentence order, and average both the results. If such an averaging policy is followed for a particular model on the dev set, we follow the same policy on the test set. The \textit{rev} subscript shall be used with dataset names to indicate that the data has been doubled using sentence reversal augmentation or with a model name to indicate that the model performs the reverse sentence averaging at inference for more robust predictions.

\begin{figure}
\hspace{-0.15cm}\includegraphics[scale = 0.32]{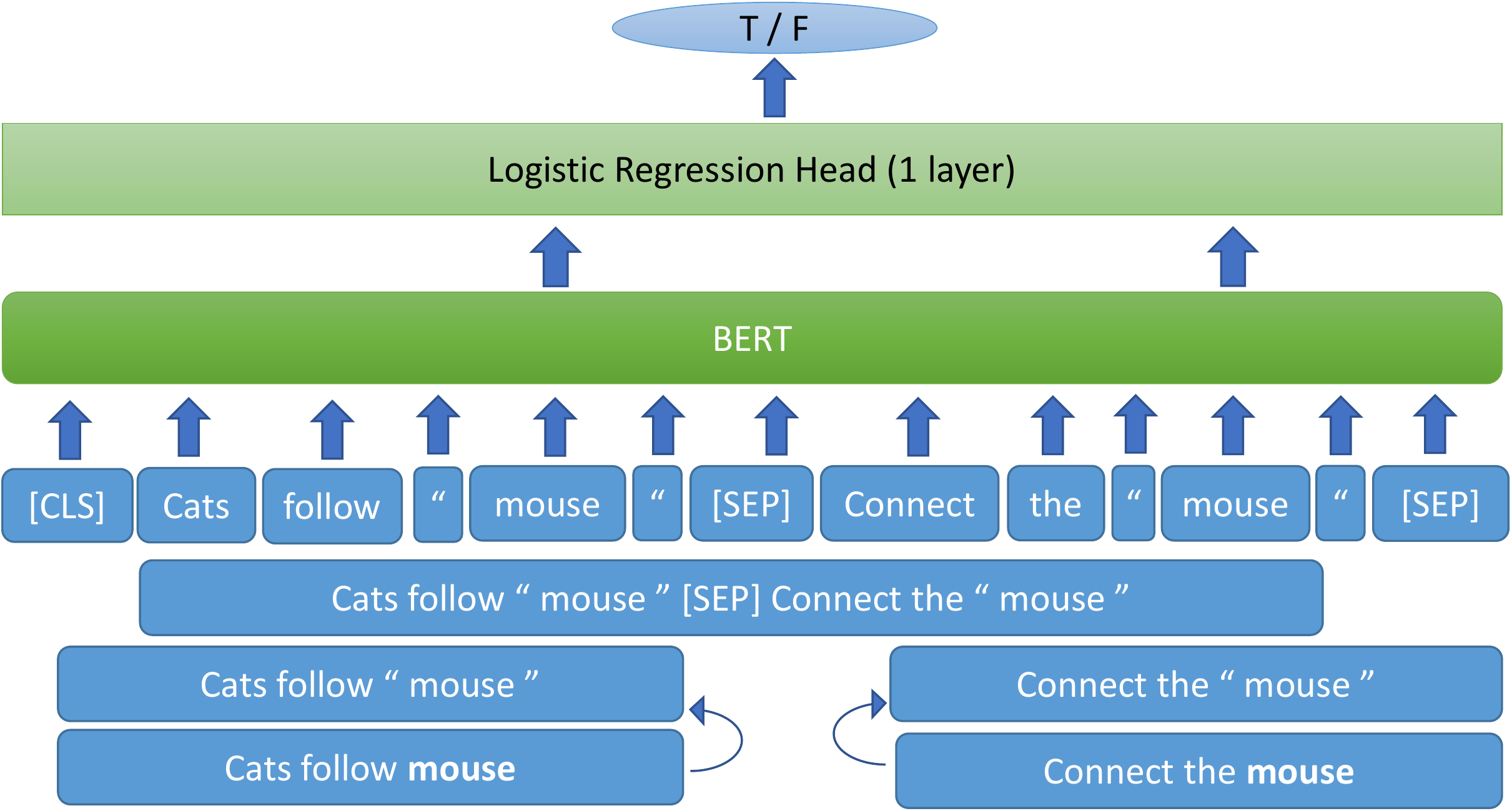}
\caption{The Transformers + Logistic Regression Architecture (+ \textit{Signal 1}). This is our proposed architecture.}
\label{fig:2}
\end{figure}

\noindent\textbf{Transformers+Logistic Regression: } Here we use the transformer-based pre-trained language model as an encoder network, feeding it with the sentence pairs concatenated with a separator token (\small$[CLS]E(x^i_{sent1} [SEP] x^i_{sent2}[SEP])$ \normalsize). We then extract the word level embeddings (last layer hidden state) for each instance of the word (from both sentence 1 and sentence 2). If the word gets sub-tokenized, we pick the embedding of the first sub-token. We finally feed their concatenation to a logistic regression head, with binary cross-entropy as the loss function. The architecture can be seen in Figure \ref{fig:2}. We used ELECTRA  \cite{clark2020electra}, ALBERT \cite{lan2019albert}, XLM-R \cite{xlmr} for English only data, and XLM-R for all other language data. Unless otherwise mentioned, we use the `\textit{large}' variant for ELECTRA and XLM-R, and the `\textit{xxlarge}' variant for ALBERT.

\noindent\textbf{Siamese Architecture: } Here, we cast our problem as a similarity problem. Similarity being measured by the closeness of the senses of the target word in each of the sentence. We, therefore, use a Transformer-based Pretrained Language Model to obtain the contextualized representations of the target word across the two sentences (using same model weights for both sentences), and optimize the contrastive loss \cite{contrastiveloss} -
	\begin{displaymath}
	    \mathcal{L} = \sum^{|T|}_{i}D_i^2 + \sum^{|F|}_j max (0, m-D_j)^2
	\end{displaymath}
Where the set $T$ is a set of same meaning sentence pairs, and $F$ has different meaning pairs. $D_i$ is the distance metric between the contextualized representations obtained for the target word by the language model for the pair of sentences. A small $D_i$ would mean that the senses of the common word across the two sentences are the same. $m$ is the margin parameter of the loss.  We experiment with L$_2$ and Cosine distances. We found this method to be good but not competitive with the Transformers+Logistic Regression method.

\subsection{English (EN-EN) Task Proposal}
For English language pair, we have dedicated training data (which we shall abbreviate as MCL-EN) that is used to train models. We used the proposed Transformers+Logistic Regression architecture and the preprocessing method of including signals using double-quotes.

In addition to sentence reversal augmentation, we augment the data using WiC and AuSemCor (Section \ref{sec:data}). Both the WiC and the AuSemCor data have been automatically created using certain heuristics (Section \ref{sec:data}), which make the sense distinctions in both the dataset a little different than what a human annotator would do. This is evidenced by the fact that human performance on the WiC benchmark is only 80\%, while it is 97\% on the Farsi dataset from the XL-WiC corpus, which was manually annotated. As this task's data is human annotated, we do not use sentence reversal augmentation with WiC and AuSemCor data to avoid an over-representation of the automatically annotated data in our training set.

\subsection{Multilingual Task (except EN-EN) Proposal}
\label{sec:multi}
For this task as well, we use Transformers+Logistic Regression architecture (the transformer being the multilingual XLM-R \cite{xlmr}), with double-quote signal preprocessing and sentence reversal augmentation.

\noindent\textbf{Data: } For the four languages under this task, we have no training data. To address that, we split the development set for each of the language pairs into a 9:1 train-dev split. The split can be done in two ways - a random split or an out-of-vocabulary split (the 1 split will primarily have words not present in the 9 split). The latter's motivation is to simulate the test set because the test set's words will be unseen (not seen during training). The former may be useful as well because the model can see and learn more words during training. This distinction is based on \citet{xlwic}'s observation that models tend to perform better on seen words than unseen words during evaluation. We experiment using both split types. All languages' data is concatenated together, and we solve the task for these four languages together by a single XLM-R model. We also use the EN-EN data of our task during this training. For development, we finally have 100 $\times$ 4 data points, which we augment again by sentence reversal augmentation, thereby finally obtaining a dev set of 800 pairs. Since we perform sentence reversal on dev set, at test inference, we follow the reverse averaging policy as described before. This is denoted by a subscript \textit{rev} in the model names. In our further discussion, we shall refer to this multilingual train data as MCL-MN$^{rand}$, for the random split method, or MCL-MN$^{oov}$ for the out-of-vocabulary split method.

We augment our data using XL-WiC (Section \ref{sec:data}). However, we do not use the XL-WiC  data on all models, and whenever we do, we do not perform sentence reversal, to prevent higher representation of lower quality data. Sentence reversal is also not performed with English to have as much proportion of non-English data as possible in the training phase.

\subsection{Cross-Lingual Task Proposal}
For this task, we have no training or development data. We propose two methods - (i) Translate-Test, (ii) Multilingual Zero-Shot.

\noindent\textbf{Translate-Test: } In this method, we use Microsoft Translator\footnote{https://azure.microsoft.com/en-us/services/cognitive-services/translator/} to translate the second sentence (in either  AR, FR, RU, ZH) of the test set to English, thereby reducing it to an EN-EN task. However, this method is bound to introduce inaccuracies from translation, and we lose positional information about the target word in the translated sentence. So we cannot use word-level embeddings , and therefore the Transformer+Logistic Regression Model (Section \ref{taskagnostic}) cannot be used. We, therefore tweak the Transformers+Logistic Regression architecture a little - we use the [CLS] token embedding instead of the word-level embeddings, keeping the Logistic regression head intact. For development, we use a back-translated EN-EN dev set (EN to FR to EN) for the second sentence to simulate inaccuracies of translation that will be induced in the second ~sentence in the test set. For training as well, we back translate 50\% of the second sentence (12.5\% to each AR, FR, RU, and ZH and back to en). Due to the loss of positional information about the target word after translation, we experiment with using \textit{Signal 2} - appending the target word at the end of the second sentence. We use ELECTRA \cite{clark2020electra} as the encoder for this sub-task.

\noindent\textbf{Multilingual Zero-Shot: } In this method, we directly use the models obtained from our Multilingual task (section \ref{sec:multi}) on the Cross-Lingual test data.

\section{Experiments and Results}
\label{sec:exp}

\begin{table*}[h!]
\centering
\begin{center}
    \begin{tabular}{|p{5cm}|c|c|c|}
    \hline
       Model & Trained On/Ensembled On & Dev  & Test \\ \hline \hline
       
       XLM-R & MCL-EN$_{rev}$ + WiC & 89.1 & 89.5 \\
       XLM-R$_{rev}$ & MCL-EN$_{rev}$ + WiC & 89.3 & 89.5 \\
       XLM-R & MCL-MN$^{oov}_{rev}$ + MCL-EN & 89.1 & 89.9 \\
       ELECTRA & MCL-EN$_{rev}$ & \textbf{91.1} & 90.3 \\
       ELECTRA$_{rev}$ & MCL-EN$_{rev}$ & 90.8 & 91.7 \\
       ELECTRA & MCL-EN$_{rev}$ + WiC + AuSemCor & 89.7 & 91.6 \\
       ELECTRA$_{rev}$ & MCL-EN$_{rev}$ + WiC + AuSemCor & 90.5  & 90.9 \\
       ALBERT & MCL-EN$_{rev}$ & 87.8 & 89.6 \\
       ALBERT$_{rev}$ & MCL-EN$_{rev}$ & 89.7 & \textbf{92.2} \\
       \hline

            Probability sum ensemble & All above models& \textbf{92.8} & \textbf{93.3}\\ 
           Majority vote ensemble  &All above models& 92.7 & \textbf{93.3}\\ 
           Probability sum ensemble & Only MCL models & 91.9 
        & 92.6\\
         
         \hline
    \end{tabular}
\end{center}
\caption{Accuracies of the models on English (EN-EN) DataSet. All are + \textit{Signal 1} models.}
\label{results:en}
\end{table*}

       

We ran various experiments to test out the efficacy of the different approaches. All experiments have been carried out with a learning rate set at $10^{-5}$, using AdamW \cite{adam} optimizer, with batch sizes varying in $\{8, 16, 32\}.$ We noticed that using a batch size of 32 was ideal. However,  limited compute availability prevented us from trying it out. We trained our models for 10 epochs. To choose our best model, we performed validation multiple times during an epoch - 5 times an epoch for the EN-EN sub-task and 4 times an epoch for all other tasks. In all tables, we report accuracy, which the task's official evaluation metric.

The results and experimental set up of various models on the English set are summarized in Table \ref{results:en}. We obtain a total of 9 models for the EN-EN sub-task, four of them being \textit{rev} variants. We submitted various ensembles, and the details are present in the lower half of the table. For the three listed ensembles, we violated our averaging policy; we do not average probability scores of the model with reverse sentence pair, even if the model was saved with such a policy on the dev set as these were overall best performing models on the dev set. At 93.3\% on the test set, our model was the \textbf{best} performing model in the EN-EN task. Also noteworthy is an ensemble of models trained only on the data provided by the task, scoring 92.6\% on the test set.

The results and experimental set up of various models on the Non-English Multilingual set are summarized in Table \ref{results:mn} for the development sets, and Table \ref{results:mn_res} for test set performance (where we show scores of various ensembles). As described in section \ref{sec:multi}, the OOV models refer to the models that were created by training on the out-of-vocabulary split method, while the RAND models refer to the models obtained by training on data created by the random split method. All models are based on XLM-R+Logistic Regression, with double quotes signal (indicated by + \textit{Signal 1}). For evaluation on the test set, we ensemble a combination of models determined by the best performance on the joint dev set and language-wise dev set.

The performance of various translate-test models is shown in Table \ref{dev:cl}. We formed an ensemble of these models and submitted it to the leaderboard, which can be seen in Table \ref{results:cl_res}.  Also, in Table \ref{results:cl_res}, we note that the best results are obtained by zero-shot application of the models trained in the multilingual sub-task to the cross-lingual sub-task.

\begin{table*}[h!]
\centering
\begin{center}
    \begin{tabular}{|c|c|c|c|c|c|c|c|c|c|c|}
    \hline
        Submission & Ensemble Details  & Dev  & \multicolumn{4}{c|}{Multilingual Test} \\ \cline{4-7}
        & & & AR & FR & RU & ZH \\\hline \hline

      1& OOV & 88 & 84.5 & 86.2 & 86.1 & 86.4\\ 
      2  & OOV$_{rev}$& 88 & 84.4 &87.5 &85.4 & 85.6 \\
      - & OOV$^2_{rev}$& 88.88 & 85.5 & \textbf{87.8} & 85.4 & 85.5\\ \hline\hline
       - & RAND$_{rev}$ & 89.38 & \textbf{86.0} & 86.6 &86.2 & 86.2\\
       - & RAND$^2_{rev}$ & 90.5 & 85.7 & 86.7 & \textbf{86.9} & 86.0 \\ \hline\hline
       - & Prob Sum (RAND$^2_{rev}$ and OOV$^2_{rev}$) & NA & 85.5 & 87.6 &86.7 & \textbf{87.3}\\
       \hline

    \end{tabular}

\caption{Final Ensembles Non-English Multilingual. A ``-" indicates model not submitted to the leaderboard.}
\label{results:mn_res}
\end{center}

\end{table*}

\begin{table*}[h!]
\begin{center}

    \begin{tabular}{|c|c|}
    \hline
        Model  & Accuracy \\ \hline \hline
       
        ELECTRA+\small\textit{Signal 2}\normalsize & \textbf{86.4}\\ 
        ELECTRA Back-T+\small\textit{Signal 2}\normalsize& 86.1\\ 
         ELECTRA Back-T$_{rev}$ & 85.6 \\ \hline

    \end{tabular}
\caption{Translate Test Models evaluated on Back Translated EN-EN dev set. Back-T Models refer to models where 50\% training data was also back translated.}
\label{dev:cl}
\end{center}
\end{table*}

\begin{table*}[h!]
\centering
\begin{center}
    \begin{tabular}{|M{3.7cm}|p{5cm}|M{0.80cm}|M{0.7cm}|M{0.7cm}|M{0.7cm}|M{0.7cm}|}
    \hline
        Model & Trained on  & Dev  & \multicolumn{4}{c|}{Language-Wise Dev} \\ \cline{4-7}
        & & & AR & FR & RU & ZH\\\hline \hline
      
       RAND XLM-R +\small\textit{Signal 1} & MCL-EN+MCL-MN$^{rand}_{rev}$& 87.38 & 89 &85 &91 & 96 \\ 
      \vspace{0.3cm} RAND XLM-R +\small\textit{Signal 1} & MCL-EN+MCL-MN$^{rand}_{rev}$+XL-WiC(ZH, DA, HR, NL)&   \vspace{0.3cm}  88.13 & \vspace{0.3cm} 88 &\vspace{0.3cm}  84.5 &\vspace{0.3cm}  91 &\vspace{0.3cm}  95.5\\   \hline\hline
      OOV XLM-R +\small\textit{Signal 1}& MCL-EN+MCL-MN$^{oov}_{rev}$& 87 & 89 & 91.5 & 91 & 83.5\\ 
      \vspace{0.3cm} OOV XLM-R +\small\textit{Signal 1} & MCL-EN+MCL-MN$^{oov}_{rev}$+XL-WiC(ZH, DA, HR, NL)& \vspace{0.3cm} 87 &\vspace{0.3cm} 87.5 &\vspace{0.3cm} 91 &\vspace{0.3cm} 92.5 &\vspace{0.3cm} 82 \\ \hline

    \end{tabular}
    \begin{tablenotes}
        \small \item Note: The development sets for RAND and OOV models are different and hence are incomparable in performance.
    \end{tablenotes}

\caption{Non-English MultiLingual Development Set Accuracies. The Dev columns indicates the performance on the joint dev set, while the Language-Wise column lists down score for that particular language's dev.}
\label{results:mn}
\end{center}

\end{table*}

\begin{table*}[h!]
\centering
\begin{center}
    \begin{tabular}{|c|c|c|c|c|c|c|c|c|c|c|}
    \hline
        Submission & Ensemble Details  & Dev  & \multicolumn{4}{c|}{Cross-Lingual Test} \\ \cline{4-7}
        & & & AR & FR & RU & ZH \\\hline \hline

      -  & OOV$_{rev}$& - & \textbf{86.9} & \textbf{87.5} &87.6 & \textbf{87.6} \\
      - & OOV$^2_{rev}$& - & 85.6 & 86.8 & 87.1 & 87.5\\ 
       - & RAND$_{rev}$ & - & 83.9 & 85.4 &86.0 & 86.1\\
       - & RAND$^2_{rev}$ & - & 85.4 & 86.7 &86.9 & 84.5 \\
       - & Prob Sum & - & 85.9 & 86.6 & \textbf{88.0} & 86.2\\ \hline
       1 & TT Ensemble 1 & 87.1 & 83.7 & 85.3 & 86.0 & 86.1\\
       - & TT Ensemble 2 & 87.3 & 83.9 & 84.8 & 85.1 & 86.5\\ \hline
       - & Adjusted Threshold RAND$_{rev}$ & - & \textbf{87.1} & \textbf{88.5} & \textbf{89} & \textbf{90.6} \\

       \hline

    \end{tabular}

\caption{Final Ensembles Non-English Cross-Lingual Test. Translate Test models have been abbreviated as TT. The Dev column is only relevant to indicate the performance of TT models on the Back Translated EN-EN dev. The adjusted threshold model is purely for the purpose of analysis (Section \ref{sec:error_analysis}).}
\label{results:cl_res}
\end{center}

\end{table*}

\section{Ablation Study}
We perform an analysis of the various proposed approaches (Table \ref{dev:en}), specifically paying heed to the EN-EN task, starting with a baseline of a BERT base model with a logistic regression head over the [CLS] token (the `cls' models in the table). We see an improvement in performance by switching to target word embeddings (the last layer hidden state corresponding to the first sub-word token). Adding the signal in the form of double quotes (\textit{Signal 1}) improves performance, probably by emphasizing the word to disambiguate, that, otherwise, the model does not know. \textit{Signal 2} however, is not as effective, but it still improves performance over the non-signal model. Utilizing sentence reversal data augmentation, we see an improvement in performance. The model's weights are not symmetric with respect to sentence 1 and 2, and ideally, we should train the model to lose its sense of order and probably make better internal representations in the process. We do not observe a significant change in model performance on using the augmented data. In fact, for BERT$_{base}$, it decays a little. Running models trained exclusively on WiC and AuSemCor, we note that the AuSemCor model performs better, but both models lag behind the model trained on MCL-EN. We note strong improvements in performance by ensembling various models (Table \ref{results:en}). The use of different transformers give us a diversity in our ensemble, with different models canceling each other's mistakes. We also observe that the ensemble of models trained using only MCL data lag behind the ensemble of models trained using the augmented data (Table \ref{results:en}). This means that the augmented data is of benefit to our models. This was not clear with just a single model, BERT$_{base}$, as mentioned before, with performance remaining around the same mark.  As mentioned before, the Siamese models (83.5\%) trail the Logistic Regression models (86.8\%). Note that for Siamese models, the sentence order is irrelevant, so we cannot perform sentence reversal augmentation, and so 83.5\% is their best with all our methods.

The analysis done on dev data is shown in Table \ref{dev:en}. We finally took the models which were giving dev accuracy greater than 89\% for the ensembles. That is XLM-R, ALBERT and ELECTRA.
Another interesting point is that an XLM-R model trained for Non-English Multilingual subtask (using MCL-MN$_{rev}$ + MCL-EN) could also slightly improve itself (at the very least, it did not degrade) than when it trained on MCL-EN$_{rev}$ data.

\begin{table}[t]
    \begin{tabular}{|p{5.58cm}|M{1.3cm}|}
    \hline
        Model  & Accuracy \\ \hline \hline
       cls BERT$_{base}$  & 83.9\\ \hline
        BERT$_{base}$  & 84.6\\ \hline
        cls BERT$_{base}$+\small\textit{Signal 1}\normalsize  & 85.3\\ \hline
        cls BERT$_{base}$+\small\textit{Signal 2}\normalsize  & 84.3\\ \hline
        BERT$_{base}$+\small\textit{Signal 1}\normalsize& 86.1\\ \hline
        BERT$_{base}$ +\small\textit{Signal 1 }\normalsize (WiC)& 72.6\\ \hline
        BERT$_{base}$ +\small\textit{Signal 1 }\normalsize (AuSemCor)& 77.5\\ \hline
        BERT$_{base}$ +\small\textit{Signal 1 }\normalsize (MCL-EN + WiC + AuSemCor)& 85.7\\ \hline
         BERT$_{base}$+\small\textit{Signal 1}\normalsize (MCL-EN$_{rev}$) & 86.8 \\ \hline

         BERT$_{Large}$+\small\textit{Signal 1 }\normalsize (MCL-EN$_{rev}$) & 87.7\\ \hline
         
        RoBERTa +\small\textit{Signal 1}\normalsize (MCL-EN$_{rev}$)& 87.7\\ \hline
         
         XLM-R +\small\textit{Signal 1}\normalsize (MCL-EN$_{rev}$ + WiC)& 89.1\\ \hline
         XLM-R +\small\textit{Signal 1 }\normalsize (MCL-EN + MCL-MN$^{rand}_{rev}$)
        & 89.3\\ \hline
        ALBERT +\small\textit{Signal 1 }\normalsize (MCL-EN$_{rev}$)
        & 89.8\\ \hline
        ELECTRA +\small\textit{Signal 1 }\normalsize (MCL-EN$_{rev}$) & \textbf{91.1}\\
         \hline
         Sia. BERT$_{base}$ + \small\textit{Signal 1 }\normalsize, L$_2$ dist. &81.8 \\ \hline
         Sia. BERT$_{base}$ + \small\textit{Signal 1 }\normalsize, Cosine 
         dist.  &83.5 \\ \hline
    \end{tabular}
\caption{An analysis of performance of models on EN-EN Dev. The training data was MCL-EN unless otherwise specified.}
\label{dev:en}
\end{table}

For the Multilingual setting, we note that there is not much difference between the performance of OOV and RAND models (Table \ref{results:mn_res}). The OOV method works better for FR, while RAND works better for AR, RU, and ZH. On average, RAND scores are slightly better, but not by a large margin.

We note that some pre-trained language models perform better for this task, especially ELECTRA and ALBERT, which improve upon the scores of RoBERTa \cite{liu2019roberta}, and BERT. We also note that their `large' variants are always better performers than the `base' variants.

\begin{table}[t]
\centering
\begin{center}
    \begin{tabular}{|l|p{0.8cm}|p{0.7cm}|p{0.8cm}|}
    \hline
    \diagbox{Actual}{Predicted} & True & False \\ \hline
      True & 478 & 22  \\ \hline
      False & 45 & 455 \\ \hline
      
    \end{tabular}
    
\end{center}
\caption{Confusion Matrix of English (EN-EN) Sub-Task}
\label{enerr}
\end{table}

\begin{table}[t]
\centering
\begin{center}
    \begin{tabular}{|l|p{0.8cm}|p{0.7cm}|p{0.8cm}|}
    \hline
    \diagbox{Actual}{Predicted} & True & False \\ \hline
      True & 1773 & 227  \\ \hline
      False & 344 & 1656 \\ \hline
      
    \end{tabular}
    
\end{center}
\caption{Confusion Matrix of Multi-Lingual (Non-English) Sub-Task (Model OOV$_{rev}$).}
\label{mnerr}
\end{table}

\begin{table}[t]
\centering
\begin{center}
    \begin{tabular}{|l|p{0.8cm}|p{0.7cm}|p{0.8cm}|}
    \hline
    \diagbox{Actual}{Predicted} & True & False \\ \hline
      True & 1627 & 373  \\ \hline
      False & 131 & 1869 \\ \hline
      
    \end{tabular}
    
\end{center}
\caption{Confusion Matrix of Zero-Shot model in Cross-Lingual Sub-Task (Model RAND$_{rev}$).}
\label{clerrz}
\end{table}

\begin{table}[h]
\centering
\begin{center}
    \begin{tabular}{|l|p{0.8cm}|p{0.7cm}|p{0.8cm}|}
    \hline
    \diagbox{Actual}{Predicted} & True & False \\ \hline
      True & 1700 & 300  \\ \hline
      False & 289 & 1711 \\ \hline
      
    \end{tabular}
    
\end{center}
\caption{Confusion Matrix of Translate-Test model in Cross-Lingual Sub-Task (Model TT Ensemble 1).}
\label{clerrt}
\end{table}

\begin{table}[h!]
\centering
\begin{center}
    \begin{tabular}{|c|c|c|c|}
    \hline
    POS & Accuracy \\ \hline
      Adverb & 86.67  \\ \hline
      Adjective & 91.6 \\ \hline
      Noun & 93.37 \\ \hline
      Verb & 94.63 \\ \hline
      
    \end{tabular}
    
\end{center}
\caption{POS wise accuracy analysis.}
\label{enpos}
\end{table}

\section{Error Analysis}
\label{sec:error_analysis}
The confusion matrices for our best model are shown in Tables \ref{enerr}, \ref{mnerr}, \ref{clerrz} and \ref{clerrt}.  On the Multilingual task, we found that number of false positives is higher than the number of false negatives for all languages.

In contrast, for the Cross-Lingual task, using the multilingual models in the zero-shot setting gives significantly lower false positives than false negatives (Table \ref{clerrz}). The Translate-Test method offered a fairly balanced prediction, albeit with lower overall accuracy. (Table \ref{clerrt}).

We observed that the probability scores returned by the multilingual models, when tested on the cross-lingual dataset, fell. We also observe a much lower number of false positives on this test data than the multilingual test data. This suggests that we need to tweak the prediction threshold value, in particular, to bring it down to adjust for the lower scores on this data set. Since we do not have a dev set for the cross-lingual sub-task, we perform the analysis on the test set itself. In Table \ref{results:cl_res}, we can see a model's performance in a threshold tuning experiment, where the threshold was brought down to 0.17 from 0.5 for all language pairs (i.e. 0.17 was used commonly for all language pairs). A significant spike in performance is observed (an average rise of 3.45\% across all the cross-lingual language pairs). This suggests that models trained on multilingual data can competitively distinguish senses in the cross-lingual setting as well, provided we take into account the fall in probability scores, induced by the transfer, by threshold moving. As a control, a similar threshold tuning experiment for the same model on the multilingual test data yielded only an average of 0.075\% improvement for the four Non-English language pairs.


We also analyzed the Parts-Of-Speech (POS) wise performance of our English Model in Table \ref{enpos}. We see a lag in performance for Adverbs and Adjectives that have less number of training data points.

\vspace{-1mm}
\section{Conclusion}

In this work, we presented our approach to solving the SemEval Task 2:  Cross-Lingual and Multilingual Word in Context Disambiguation.
We proposed different models based on transformers for the English, Non-English Multilingual and Cross-Lingual tasks. The application of signals and sentence reversal augmentation helped us improve performance across all tasks. Utilising the existing SemCor dataset, we created AuSemCor for the EN-EN sub-task. Due to the unavailability of training data in the Non-English Multilingual and Cross-Lingual task, we proposed methods to obtain the training data from the dev sets or external resources.  We finally submitted ensembles for all tasks.

\bibliographystyle{acl_natbib}
\bibliography{acl2021}


\end{document}